\documentclass[letterpaper]{article}

\usepackage{uai2019}
\usepackage{times}

\usepackage[margin=1in]{geometry}

\usepackage{amsmath,amsfonts,bm}

\def\eqref#1{equation~(\ref{#1})}

\def\1{\bm{1}}

\DeclareMathAlphabet{\mathsfit}{\encodingdefault}{\sfdefault}{m}{sl}
\SetMathAlphabet{\mathsfit}{bold}{\encodingdefault}{\sfdefault}{bx}{n}

\usepackage{hyperref}
\usepackage{url}
\usepackage{natbib}   
\usepackage{graphicx}
\usepackage{gensymb}
\usepackage{amsmath}
\usepackage{enumitem}
\usepackage{amssymb}
\usepackage{fancyhdr}
\usepackage[ruled,vlined]{algorithm2e}
\usepackage{wrapfig}
\usepackage{gensymb}
\usepackage{bbm}
\usepackage[colorinlistoftodos, textsize=small]{todonotes}
\usepackage{multirow}
\usepackage[toc,page]{appendix}
\usepackage{booktabs}

\title{Differentiable probabilistic models of scientific imaging \\ with the Fourier slice theorem}

\author{ {\bf Karen Ullrich\thanks{\hspace{5pt} \texttt{mail.karen.ullrich@gmail.com}}} \\
University of Amsterdam \\
\And
{\bf Rianne van den Berg}  \\
University of Amsterdam \\
\And
{\bf Marcus Brubaker}   \\
York University\\
\And
{\bf David Fleet\thanks{\hspace{5pt} CIFAR AI Chair, Vector Institute, CIFAR LMB Program} }\\
University of Toronto \\
\And
{\bf Max Welling\thanks{\hspace{5pt} Qualcomm, CIFAR LMB Program} } \\
University of Amsterdam 
}

\begin{document}

\maketitle

\begin{abstract}
Scientific imaging techniques, e.g., optical and electron microscopy or computed tomography, are used to study 3D structures through 2D observations. 
These observations are related to the 3D object through orthogonal integral projections. 
For computational efficiency, common 3D reconstruction algorithms model 3D structures in Fourier space, 
exploiting the Fourier slice theorem.
At present it is somewhat unclear how to differentiate through the projection operator 
as required by learning algorithms with gradient-based optimization.  
This paper shows how back-propagation through the projection operator in Fourier space can be achieved. 
We demonstrate the approach on 3D protein reconstruction. 
We further extend the approach to learning probabilistic 3D object models. 
This allows us to predict regions of low sampling rates or to estimate noise. 
Higher sample efficiency can be reached by utilizing the learned uncertainties of the 3D structure as an unsupervised estimate of model fit. 
Finally, we demonstrate how the reconstruction algorithm can be extended with amortized inference on unknown attributes such as object pose. Empirical studies show that joint inference of the 3D structure and object pose becomes difficult when the underlying object contains symmetries, in which case pose estimation can easily get stuck in local optima, inhibiting a fine-grained high-quality estimate of the 3D structure.  
\end{abstract}
\begin{figure}[!bht]
  \begin{center}
    \includegraphics[width=0.45\textwidth]{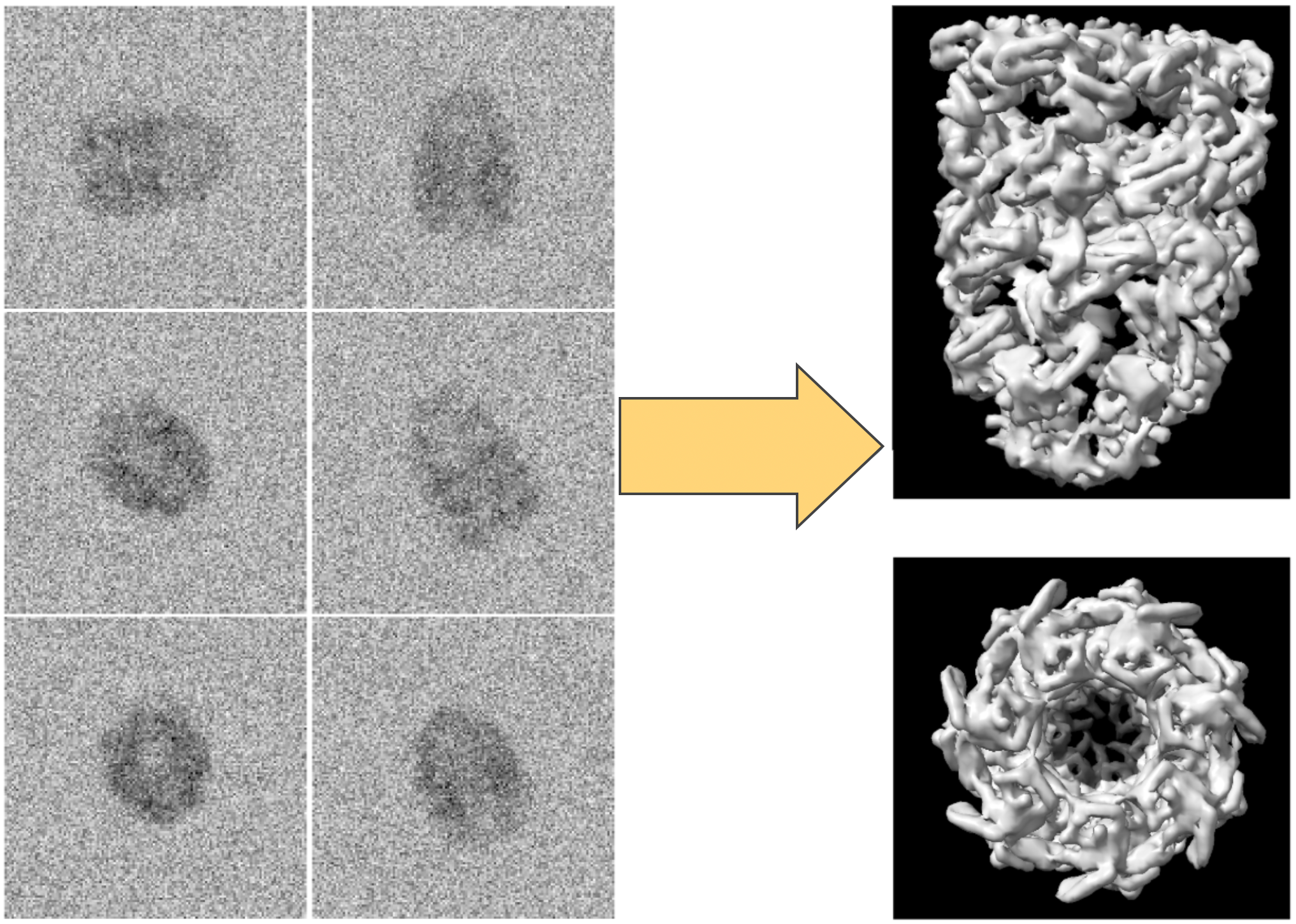}
  \end{center}
  \caption{Example of electron cryo-microscopy with the GroEL-GroES protein \citep{xu1997crystal}. 
  \textit{Left}: 2D observations obtained by projections with an electron beam. \textit{Right}: Two different views of the ground truth 3D protein structure represented by its electron density.}
\label{fig:coneuncertainties}
\end{figure}

\section{Introduction}\label{sec:introduction}
The main goal of many scientific imaging methods is to reconstruct a $(d+1)$-dimensional structure $\textbf{v} \in \mathcal{V}\subseteq\mathbb{R}^{D^{d+1}}$ from $N$ (d)-dimensional observations $\textbf{x}_n \in \mathcal{I} \subseteq\mathbb{R}^{D^{d}}$, where $d$ is either one or two. For the sake of simplicity we will talk about the case $d=2$ in the rest of this work.
The contributions of this paper are:
\begin{enumerate}
    \itemsep 0.0in
    \item We view the process of image formation through a graphical model in which latent variables correspond to physical quantities such as the hidden structure $\textbf{v}$ or the relative orientation/pose of a specimen. This enables one to predict errors in the reconstruction of 3D structures through uncertainty estimates. This is especially interesting when objects $\textbf{v}$ are only partially observable, as is the case in certain medical scans, such as breast cancer scans. Moreover, uncertainty prediction enables more data efficient model validation.
    \item Based on the aforementioned innovations, we propose a new method for (unsupervised) reconstruction evaluation. Particularly, we demonstrate that learned uncertainties can replace currently used data inefficient methods of evaluation (see Section \ref{sec:exp_data_eff}). 
    We thus learn better model fits than traditional methods given the same amount of data.
    \item We extend current approaches such as \citep{jaitly2010bayesian} to describe the generative process as a differentiable map by adopting recent techniques from the deep learning community \citep{jaderberg2015spatial, rezende2016unsupervised}. We demonstrate that this fenables more advanced joint inference schemes over object pose and structure estimation.    
\end{enumerate}

Our experimental validation focuses on single particle electron cryo-microscopy (cryoEM). 
CryoEM is a challenging scientific imaging task, as it suffers from complex sources of observation noise, 
low signal to noise ratios, and interference corruption.
Interference corruption attenuates certain Fourier frequencies in the observations. Radiation exposure is minimized because electron radiation severely damages biological specimens during data collection. Minimal radiation, however, leads to low signal-to-noise ratios, where sensor cells record relatively low electron counts.
Since imaging techniques like CT suffer from a subset of these difficulties, we believe that evaluating and analyzing our method on cryoEM problems is appropriate.

\section{Background and related work}

Modelling nano-scale structures such as proteins or viruses is a central task in structural biology. 
By freezing such structures and subsequently projecting them via a parallel electron beam to a sensor grid (see figure \ref{fig:image_formation}), CryoEM enables reconstruction and visualization of such structures. The technique has been described as revolutionary because researchers are capable of observing structures that cannot be crystallized, as required for X-ray crystallography \citep{rupp2009biomolecular}.

The main task of reconstructing the structure from projections in cryoEM, and the wider field of medical imaging, is somewhat similar to multi-view scene reconstruction from natural images. 
There are, however, substantial differences.  Most significantly, the projection operation 
in medical imaging is often an orthogonal integral projection, while in computer vision it
is a non-linear perspective projection for which materials exhibit different degrees of opacity. 
Thus, the generative model in computer vision is more complex. 
Medical imaging domains, on the other hand, face significant noise and measurement uncertainties, with signal-to-noise ratios as low as 0.05 \citep{baxter2009determination}.

Most CryoEM techniques \citep{de2013xmipp,grigorieff2007frealign,scheres2012relion,tang2007eman2} iteratively 
refine an initial structure by matching a maximum a posteriori (MAP) estimate of the pose (orientation and position) under the proposal structure with the image observation. These approaches suffer from a range of problems such as high sensitivity to poor initialization \citep{henderson2012outcome}.
In contrast to this approach, and  closely related to our work, \citep{brubaker2015building,punjani2017cryosparc} 
treat poses and structure as latent variables of a joint density model. 
MAP estimation enables efficient optimization in observation space.   
Previous work \citep{sigworth1998maximum, scheres2007disentangling,jaitly2010bayesian,scheres2012relion} 
has suggested full marginalization, however due to its cost, it is usually intractable. 

This paper extends the MAP approach by utilizing variational inference to approximate intractable integrals. 
Further, reparameterizing posterior distributions enables gradient based learning \citep{kingma2013auto,rezende2014stochastic}. 
To our knowledge this is the first such approach that provides an efficient way to learn approximate posterior distributions in this domain.

\section{Observation Formation Model}\label{sec:image_formation}

Given a structure $\textbf v$, we consider a generic generative model of observations, 
one that is common to many imaging modalities.
As a specific example, we take the structure $\textbf v$ to be a frozen (i.e. cryogenic) protein complex, although the procedure described below applies as well to CT scanning and optical microscopy. 
$\textbf v$ is in a specific pose $\textbf{p}_n$ relative to the direction of the 
electron radiation beam.  This yields a pose-conditional projection, with observation $\textbf{x}_n$.
Specifically, the pose $\textbf{p}_n =(\textbf{r}_n,\textbf{t}_n)$, consists 
of $\textbf{r}_n \in SO(3)$, corresponding to the rotation of the object 
with respect to the microscope coordinate frame, and $\textbf{t}_n \in E(3)$, the 
translation of the structure with respect to the origin.

\begin{figure}[!t]
  \begin{center}
    \includegraphics[scale=0.35]{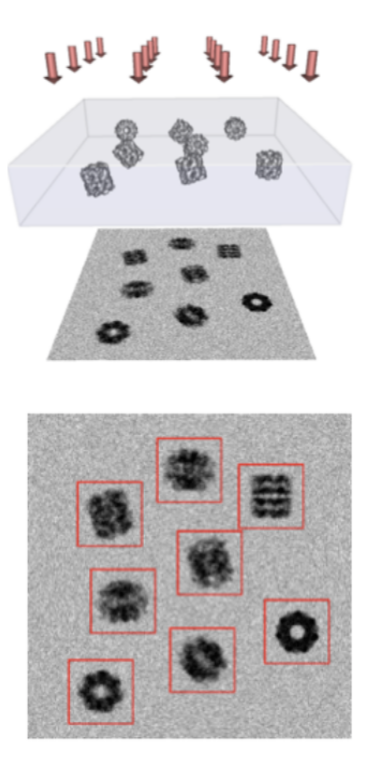}
  \end{center}
  \caption{\textit{Top:} Image formation on the example of cryo EM: The parallel Electron beam projects the electron densities on a surface where a grid of DDD sensors record the number of electrons that hit it. \textit{Bottom:} To detect the projections (outlines in red) an algorithm  seeks out areas of interest \citep{langlois2014automated,zhao2013tmacs} (Figure from \citep{Pintilie}).}
\label{fig:image_formation}
\end{figure}

The observations are then generated as follows:
The specimen in pose $\textbf p_n$ is subjected to radiation (the electron beam), 
yielding an orthographic integral projection onto the plane perpendicular to the beam.
The expected projection can be formulated as
\begin{align}
\bar{\textbf{x}}_n = P\, T_{\textbf{t}_n} R_{\textbf{r}_n} \mathbf{v}\,.
\label{eq:expected_projection}
\end{align}
Here $P$ is the projection operator, $T_{\textbf{t}_n}$ is the linear translation operator 
for translation $\textbf{t}_n$, and $R_{\textbf{r}_n}$ is the linear operator corresponding to rotation $\textbf{r}_n$. 
Without loss of generality we can choose the projection direction to be along the $z$-direction $\textbf{e}_z$.
When the projection is recorded with a discrete sensor grid (i.e., sampled), information beyond 
the Nyquist frequency is aliased. 
Additionally, the recording is corrupted with noise stemming from
the stochastic nature of electron detection events and sensor failures \citep{egelman2016current}. 
Low doses are necessary since electron exposure causes damage to sensitive biological molecules. 
Logically, the effect is more severe for smaller objects of study. 

Many sophisticated noise models have been proposed for these 
phenomena \citep{faruqi2003evaluation,vulovic2013image,scheres2007modeling}. 
In this work, for simplicity, we assume isotropic Gaussian noise; i.e., 
\begin{align}
 p(\textbf{x}_n  |  \textbf{p}_n, \mathbf{v} )= \mathcal{N}(\textbf{x}_n | \bar{\textbf{x}}_n, \mathbbm{1} \sigma_\epsilon^2),
\label{eq:imageformation}
\end{align}
where $\sigma_\epsilon$ models the magnitude of the observation noise.
The image formation process is depicted in Figure \ref{fig:image_formation}. 

The final section below discusses how one can generalize to more sophisticated 
(learnable) noise models.
Note that we do not model interference patterns caused by electron scattering, called defocus and modelled with a contrast transfer function (CTF). This will lead to less realistic generative models, however we see the problem of CTF estimation as somewhat independent of our problem. Ideally, we would like to model the CTF across multiple datasets, but we leave this to future work.

\section{Back-propagating through the generative model}\label{sec:fourier_slice}
In this section we aim to bridge the gap between our knowledge of the generative process $p(\textbf{x}_n|\textbf{p}_n,\textbf{v})$ 
and a differentiable mapping that facilitates direct optimization of hidden variables ($\textbf{p}_n,\textbf{v}$) with gradient-descent style schemes.
We start with an explanation of a naive differentiable implementation in position space, followed by a computationally more efficient version by shifting the computations to the Fourier domain (momentum space).

\subsection{Naive implementation: project in position space}\label{sec:naive_implemntaiton}

Our goal is to optimize the conditional log-likehood $\log p(\textbf{x}_n|\textbf{p}_n,\textbf{v})$ with respect to the unobserved $\textbf{p}_n$ and $\textbf{v}$, maximizing the likelihood of the 2D observations.
This requires \eqref{eq:imageformation} to be a differentiable operator with respect to $\textbf{p}_n$ and $\textbf{v}$. 
Note that the dependence on $\textbf{p}_n$ and $\textbf{v}$ is fully determined by \eqref{eq:expected_projection}. 
In order to achieve this, we first need to apply the group action $R_{\textbf{r}_n}$  onto $\textbf{v}$. 
Matrix representations of the group action such as the Euler angles matrix are  defined on the sampling grid 
$G = \{(\nu_1^{(j)},\nu_2^{(j)},\nu_3^{(j)})\}_{j=1}^{D^3}$ of $\textbf{v}$ rather than the voxel values $\{v_j\}_{j=1}^{D^3}$ . 
For example, the action induced by a rotation around the z-axis by an angle $\alpha$ on the position $(\nu_1^{(j)},\nu_2^{(j)},\nu_3^{(j)})$ of an arbitrary voxel $j$  can be written as,
\begin{equation}
R_\alpha \cdot \mathbf{\nu}^{(j)} =
\begin{pmatrix}
   \cos(\alpha)     &  -\sin(\alpha)   &  0\\
    \sin(\alpha)     &  \cos(\alpha)   &  0\\
    0     &  0   &  1
\end{pmatrix}
\begin{pmatrix}
   \nu_1^{(j)}\\
    \nu_2^{(j)}\\
   \nu_3^{(j)}
\end{pmatrix}.
\end{equation}
This entails two problems. First, the volume after transformation should be sampled at the same grid points as before. 
This requires interpolation. Second, to achieve a differentiable map we need to formulate the transformation of position values as a transformation of the voxel values.
\cite{jaderberg2015spatial} offers a solution to both problems, known as differentiable sampling\footnote{Originally invented to learn affine transformations on images to ease the classification task for standard neural networks, the approach has since been extend to 3D reconstruction problems from images \citep{rezende2016unsupervised}.}.

The $j$-th voxel $v^\prime_j  = ( \textbf{v}^\prime )_j$ of the transformed volume, 
$\textbf{v}^\prime = R_{\textbf{r}_n}\textbf{v}$, 
with index vector $\zeta^{(j)}$, can be expressed as a weighted sum of all voxels before 
transformation $\{v_i,\nu^{(i)}\}_{i=1}^{D^3}$.  
The weights are determined by a sampling kernel $k(\cdot)$, the argument of which is 
the difference between the transformed voxel's position $\zeta^{(j)}$ and all transformed sampling 
grid vectors $R_\alpha\cdot\nu^{(i)}$:
\begin{align}
\begin{split}
    v^\prime_j = \sum_{i=1}^{D^3} &v_i \cdot k(R_\alpha^{-1}\cdot\zeta^{(j)}-\nu^{(i)})
\end{split}
    \label{eq:interpolation}
\end{align}
A popular kernel in this context is the linear interpolation  sampling kernel\footnote{Linear kernels 
are efficient  and yield fairly good results. 
More complex ones such as the Lanczos re-sampling kernel may actually yield worse results due to smoothing.}
{\small
\begin{align}
    &k(R_\alpha^{-1}\cdot\zeta^{(j)}-\nu^{(i)}) =\notag \\
    &\prod\limits_{m=1}^3\max(0, 1- |(R_\alpha^{-1}\cdot\zeta^{(j)})_m-\nu^{(i)}_m|).
\end{align}}

Computing one voxel $v^\prime_j$  only requires a sum over 8 voxels from the original structure.
These are determined by flooring and ceiling the elements of $(R_\alpha^{-1} \zeta^{(j)})_m$. 
Furthermore, the partial derivatives are provided by, 
{\small
\begin{align}
\frac{\partial v^\prime_j}{\partial v_i} &=   k(R_\alpha^{-1}\cdot\zeta^{(j)}-\nu^{(i)})\\
\frac{\partial v^\prime_j}{\partial (R_\alpha^{-1}\zeta^{(j)})_m} &=  
\sum_{i=1}^{D^3} v_i \prod\limits_{l\neq k}\max(0, 1- |(R_\alpha^{-1}\zeta^{(j)})_l-\nu_l^{(i)}|) 
\notag\\
&\cdot
\begin{cases}
0 &\text{if } |(R_\alpha^{-1}\zeta^{(j)})_m-\nu_m^{(i)}| \geq 1\\
-1&\text{elif } (R_\alpha^{-1}\zeta^{(j)})_m \geq \nu_m^{(i)}\\
1&\text{elif }(R_\alpha^{-1}\zeta^{(j)})_m <\nu_m^{(i)}
\end{cases} .
\end{align}}

This framework was originally proposed for any differentiable kernel and any differentiable affine position transformation $\nu \rightarrow \zeta$. In our setting, we restrict ourselves to linear interpolation kernels. 
The group actions represented by $R_{\textbf{r}}$ are affine. 
In this work we represent rotations as Euler angles by using the Z-Y-Z convention.  
One could also use quaternions or exponential maps. 
As with rotation, the translation operation is also a transformation of the voxel grid, rather than the voxel values. Thus, \eqref{eq:interpolation} can also be used to obtain a differentiable translation operation.

Finally, the orthogonal integral projection operator is applied by summing voxel values along one principal direction. 
Since the position of the hidden volume is arbitrary we can fix this direction to be the Z-axis as discussed in section \ref{sec:image_formation}. 
Denoting a volume, rotated and translated according to $\textbf p = (\textbf r, \textbf t)$, by $\textbf v' = T_{\textbf t}R_{\textbf r} \textbf v$, 
the $(\zeta_1,\zeta_2)$-th element of its expected projection is given by 
\begin{align}\label{eq:projection}
    \bar{\textbf{x}}_n[\zeta_1,\zeta_2] = (P_{3\rightarrow 2}\textbf{v}')[\zeta_1,\zeta_2] = \sum_{\zeta_3} \textbf v'[\zeta_1,\zeta_2,\zeta_3],
\end{align}
where $\textbf{v}'[\zeta_1,\zeta_2,\zeta_3]$ denotes the element $(\zeta_1,\zeta_2,\zeta_3)$ of $\textbf v'$. 

This concludes a naive approach to modelling a differential map of the  expected observation in position space. 
This approach is not particularly efficient, as according to \eqref{eq:projection} we need to interpolate all $D^3$ voxels to compute one $D^2$ dimensional observation. Moreover, back-propagating through this mapping requires transporting gradients through all voxels. 
Next, we show how to reduce the cost of this naive approach without a loss of precision by shifting the problem to the Fourier domain.

\subsection{Projection-slice theorem}
The projection-slice theorem or Fourier-slice theorem states the equivalence between
the Fourier transform $F_d$ of the projection $P_{d^\prime\rightarrow d}$ of a $d^\prime$ dimensional function $f(r)$ onto a $d$-dimensional submanifold  $\mathcal{F}_{d}P_{d^\prime\rightarrow d}f(r)$
and a $d$-dimensional slice of the $d^\prime$-dimensional Fourier transform of that function. 
This slice is a $d$-dimensional linear submanifold through the origin in the Fourier domain that is parallel to the projection submanifold $\mathcal{S}_{d}\mathcal{F}_{d^\prime}f(r)$.

In our setting, given an axis-aligned projection direction (see \eqref{eq:expected_projection}) and the discrete grid $G$, 
the expected observation in 2D Fourier space is equal to a central slice through the Fourier transformed 3D structure parallel to the projection direction $\textbf{e}_z$:
\begin{align}\label{eq:slice}
\mathcal{F}_{2}\bar{\mathbf{x}}
=&\mathcal{F}_{2}P_{3\rightarrow 2}\mathbf{v}' \notag\\
=& \mathcal{S}_{2}(\mathcal{F}_{3}\textbf{v}') = \mathcal{S}_{2}\hat{\textbf{v}}',
\end{align}
where $\mathcal{F}_2$ and $\mathcal{F}_3$ denote discrete Fourier transforms in 2 and 3 dimensions. The slice operator $\mathcal{S}$
is the Fourier equivariant of the projection operator. In our case, it is applied as follows:
\begin{align}
    (\mathcal{S}_{2}\hat{\textbf{v}}')[\omega_1,\omega_2] =  \hat{\textbf{v}}'[\omega_1,\omega_2,0],
\end{align}
where $\omega_m$ are  Fourier indexes. 

This allows one to execute the generative model in position or momentum space. 
It has proven more efficient for most reconstruction algorithms to do computation in the Fourier domain \citep{sigworth2016principles}. 
This also applies to our algorithm: (i) We reconstruct the structure in the Fourier domain. 
This means we only need to apply an inverse Fourier transformation at the end of optimization. (ii) We may save the Fourier transformed expected projections a-priori, further this is easily parallelized. 
Thus even though in its original formulation we can not expect a computational benefit, when sharing the computation across many data points to reconstruct one latent structure the gain is significant. 
We elaborate on this point with respect to differentiation below.

\subsection{Differentiable orthographic integral projection}
We incorporate the Fourier Slice Theorem into our approach to build an efficient differentiable generative model. 
For this we translate all elements of the model to their counterparts in the Fourier domain. 
The Gaussian noise model becomes \citep{uncertainty_principle},
\begin{align}
     p(\mathcal{F}_2\textbf{x}_n  |  \textbf{p}_n, \hat{\textbf{v}} )= \mathcal{N}(\mathcal{F}_2\textbf{x}_n | \mathcal{F}_2\bar{\textbf{x}}_n, \mathbbm{1} \tfrac{\sigma_\epsilon^2}{2}),
\end{align}
where $\hat{\textbf{v}}=\mathcal{F}_{3}\mathbf{v}$. In the following, we aim to determine an expression for the expected Fourier projection $\mathcal{F}_2\bar{\textbf{x}}_n = \mathcal{F}_2(P_{3\rightarrow 2}T_{\textbf{t}_n}R_{\textbf{r}_n}\textbf{v})$ by deriving the operators counterparts
$ \mathcal{F}_2(P_{3\rightarrow 2}T_{\textbf{t}_n}R_{\textbf{r}_n}\textbf{v}) = \hat{P}_{3\rightarrow 2}\hat{T}_{\textbf{t}_n}\hat{R}_{\textbf{r}_n}\hat{\textbf{v}}$.

We start by noting that it is useful to keep $\hat{\textbf{v}}$ in memory $\hat{\textbf{v}}$ to avoid 
computing the 3D discrete Fourier transform multiple times during optimization. 
The inverse Fourier transform $F_{3}^{-1}\hat{\mathbf{v}}=\mathbf{v}$ is then only applied once, after convergence of the algorithm. 
Next, we restate that the Fourier transformation is a rotation-equivariant mapping $\hat{R_{\textbf{r}_n}} = R_{\textbf{r}_n}$ \citep{chirikjian2000engineering}. This means the derivations from Section \ref{sec:naive_implemntaiton} with respect to the rotation apply in this context as well. 
A translation in the Fourier domain $\hat{T}_{\textbf{t}_n}$, however, induces a re-weighting of the original Fourier coefficients,
\begin{align}\label{eq:ftranslation}
     (\hat{T}_{\textbf{t}_n}R_{\textbf{r}_n}\hat{\textbf{v}})[\omega_1,\omega_2,\omega_3] = e ^{-\textit{i} 2\pi \textbf{t}_n \cdot \omega} (R_{\textbf{r}_n}\hat{\textbf{v}})[\omega_1,\omega_2,\omega_3].
\end{align}
Finally, the last section established the equivalence of the slice and the projection operator $\hat{P}_{3\rightarrow 2}=\mathcal{S}_{2}$ (see \eqref{eq:slice}) in momentum and position space.
Specifically, for the linear interpolation kernel, we compute the set of interpolation points 
by flooring and ceiling the elements of the vector $R^{-1}_{\textbf{r}_n}\omega^{(j)}$, $\forall \omega^{(j)}=(\omega^{(j)}_1,\omega^{(j)}_2,0)$. 
This entails 6 interpolation candidates per voxel of the central slice, in total $6D^2$. 
Remember, this computation above involved $D^3$ voxels and $6D^3$ candidates. 

We can further improve the efficiency of the algorithm by swapping the projection and translation operators. 
That is, due to parallel radiation, and hence orthographic projection, 
\begin{align}
    PT_{\textbf{t}_n} R_{\textbf{r}_n} \textbf{v} = T_{\boldsymbol\tau_n}P R_{\textbf{r}_n} \textbf{v},
\end{align}
where $\boldsymbol{\tau}_n = (\textbf{t}_n \cdot \textbf e_x) \textbf e_x + (\textbf{t}_n \cdot \textbf e_y)\textbf e_y$.  This is more efficient because we reduce the translation to a two dimensional translation. Thus this modifies \eqref{eq:ftranslation} to its two dimensioanl equivalent.

For the naive implementation the cost of a real space forward projection is O$(D^3)$.  
In contrast, converting the volume to the Fourier space O$(D^3 \log D)$, 
projecting O$(D^2)$ and applying the inverse Fourier transform O$(D^2 \log D)$. 
At first glance this implies a higher computational cost. 
However, for large datasets the cost of transforming the volume is amortized over all data points.  
For gradient descent schemes, we iterate over the dataset more than once, hence the cost of Fourier transforming the observations is further amortized.
Furthermore, it is often reasonable to consider only a subset of Fourier frequencies, so back projection becomes O$(r^2)$ with $r < D$. 
The efficiency of this algorithm in the context of cryo-EM was first recognized by \citet{grigorieff1998three}. 
(We provide code for the differentiable observation model and in particular for the Fourier operators: \url{https://github.com/KarenUllrich/pytorch-backprojection}.)

\section{Variational inference}\label{sec:VI}
Here we describe the variational inference procedure for 3D reconstruction in Fourier space, enabled by the Fourier slice theorem. 
We assume we have a dataset of observations from a single type of protein with ground truth structure $\textbf v$, and its
Fourier transform $\hat{\textbf v} = \mathcal{F}_3\textbf v$. 
We consider two scenarios. 
In the first, both the pose of the protein and the projection are observed, and inference is only performed over the global protein structure. 
In the second, the pose of the protein for each observation is unknown. Therefore, inference is done over poses and the protein structure. 

The first scenario is synonymous with the setting in tomography, where we observe a frozen cell or larger complex positioned in known poses. This case is often challenging because the sample cannot be observed from all viewing angles.
For example, in cryo-EM tomography the specimen frozen in an ice slice can only be rotated till the verge of the slice comes into view. We find similar problems in CT scanning,  for example in breast cancer scans.
The second scenario is relevant for cryo-EM single particle analysis. 
In this case multiple identical particles are confined in a frozen sample and no information on their structure or position is available a priori.

In this work we lay the foundations for doing inference in either of the two scenarios. However,
our experiments  demonstrate that joint inference over the poses and the 3D structure is very sensitive to getting stuck in local minima that correspond to approximate symmetries of the 3D structure. Therefore, the main focus of this work is the setting where the poses are observed.

\begin{figure}[t]
  \begin{center}
    \includegraphics[scale=0.17]{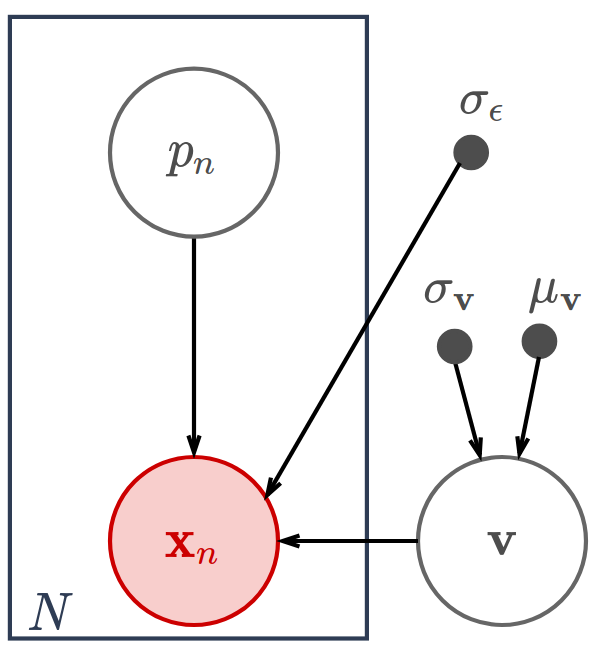}
  \end{center}
  \vspace*{-0.2cm}
  \caption{\textit{Graphical model}: Latent structure $\mathbf{v}$, pose $p_n$ and noise $\sigma_\epsilon$ can be learned from observations $\mathbf{x}_n$ through back-propagation. The latent structure distribution is thereby characterized by a set of parameters, in the Gaussian example $\mu_\mathbf{v}$ and $\sigma_\mathbf{v}$.}
\label{fig:coneuncertainties}
\end{figure}

\subsection{Inference over the 3D structure}\label{sec:VIstructure}

Here the data comprise image projections and poses, $\{(\textbf x_n, \textbf p_n)\}_{n=1}^N$,
with Fourier transformed projections denoted $\hat{\textbf x}_n = \mathcal{F}_2 \textbf x_n$.
Our goal is to learn a model $q(\hat{\textbf v})$ of the latent structure that as closely as possible resembles the true posterior $p(\hat{\textbf v}|\{\hat{\textbf x}_n\}, \{\textbf p_n\})$.
For this, we  assume a joint latent variable model $p(\{\hat{\textbf x}_n\}_{n=1}^N, \hat{\textbf v} | \{\textbf p_n\}_{n=1}^N) =p(\{\hat{\textbf x}_n\}_{n=1}^N | \{\textbf p_n\}_{n=1}^N, \hat{\textbf v}) p(\hat{\textbf v})$. 
To avoid clutter below, we use short-hand notations like $\{\hat{\textbf x}_n\}$ for $\{\hat{\textbf x}_n\}_{n=1}^N$.

Specifically, we minimize an upper bound to the Kullback-Leibler (KL) divergence:
{\small
\begin{align}
     & D_{\mathrm{KL}}\left[
     q(\hat{\textbf v})\|p(\hat{\textbf v}|\{\hat{\textbf x}_n\}, \{\textbf p_n\})
     \right] \notag \\
     &\geq - \int \mathrm d\hat{\textbf v} \; q(\hat{\textbf v}) \ln \left( \frac{p(\{\hat{\textbf x}_n\} | \{\textbf p_n\}, \hat{\textbf v}) p(\hat{\textbf v})}{q(\hat{\textbf v})}\right) \notag \\
     &= \sum_{n=1}^N - \mathbb E_{q(\hat{\textbf v})}\left[\ln p(\hat{\textbf x}_n | \textbf p_n, \hat{\textbf v}) \right] 
   + D_{\mathrm{KL}}\left[q(\hat{\textbf v})\|p(\hat{\textbf v})\right].
   \label{eq:elbo_volume}
\end{align}
}
Here, we have assumed that, given the volume, the data are IID: $p(\{\hat{\textbf x}_n\} | \{\textbf p_n\}, \hat{\textbf v}) = \prod_{n=1}^N p(\hat{\textbf x}_n | \textbf p_n, \hat{\textbf v})$. 
We have bounded the divergence by the data log-likelihood $\ln p(\{\hat{\textbf x}_n\})$, a constant with respect to $\hat{\textbf v}$ and $\{\textbf p_n\}$. 
This is equivalent to lower bounding the model evidence by introducing a variational posterior \citep{jordan1998introduction}.

In this work we focus on modelling $q(\hat{\textbf v})$ as isotropic Gaussian distribution. 
The prior is assumed to be a standard Gaussian. 
In practice, we use stochastic gradient descent-like optimization, with the data organized in mini-batches. 
That is, we learn the distribution parameters $\eta=\{\mu_\textbf{v}, \sigma_\textbf{v}\}$ for $q(\hat{\textbf v})=q_\eta(\hat{\textbf v})$ through stochastic optimization, efficiently by using the reparameterization trick \citep{kingma2013auto, rezende2014stochastic}.

In \eqref{eq:elbo_volume}, the reconstruction term depends on the number of datapoints.
The KL-divergence between the prior and approximate posterior does not. 
As the sum of the mini-batch objectives should be equal to the objective of the entire dataset, 
the mini-batch objective is 
{\small
\begin{align}
   \sum_{n \in D_i} - \mathbb E_{q(\hat{\textbf v})}\left[\ln p(\hat{\textbf x}_n | \textbf p_n, \hat{\textbf v}) \right] 
   + \frac{|D_i|}{N}D_{\mathrm{KL}}\left[q(\hat{\textbf v})\|p(\hat{\textbf v})\right]. 
   \label{eq:elbo_volume_minibatch}
\end{align}}
where $D_i$ is the set of indices of the data in minibatch $i$, and $|D_i|$ denotes the size of the $i$-th minibatch.

\subsection{Joint inference over the 3D structure and poses}

In the second scenario the pose of the 3D structure for each observation is unknown.
The data thus comprises the observed projections $\{\textbf x_n\}_{n=1}^N$. 
Again, we perform inference in the Fourier domain, with transformed projections $\{\hat{\textbf x}_n\}_{n=1}^N$ as data.
We perform joint inference over the poses $\{\textbf p_n\}_{n=1}^N$ and the volume $\hat{\textbf v}$. 
We assume the latent variable model can be factored as follows, $p(\{\hat{\textbf x}_n\}, \hat{\textbf v} , \{\textbf p_n\}) =p(\{\hat{\textbf x}_n\} | \{\textbf p_n\}, \hat{\textbf v}) p(\{\textbf p_n\}) p(\hat{\textbf v})$. 

Upper bounding the KL-divergence, as above, we obtain
{\small
\begin{align}
     &D_{\mathrm{KL}}\left[q(\hat{\textbf v})q(\{\hat{\textbf x}_n\})\|p(\hat{\textbf v},\{\hat{\textbf x}_n\}, \{\textbf p_n\})\right] \notag\\
     &\geq - \int \!\!\!\!\int \mathrm d^N \textbf p_n \mathrm d\hat{\textbf v} \; q(\{\textbf p_n\}) q(\hat{\textbf v})  \notag \\
     & \qquad \qquad \qquad \times  \ln \left( \frac{p(\{\hat{\textbf x}_n\} | \{\textbf p_n\}, \hat{\textbf v})  p(\{\textbf p_n\}) p(\hat{\textbf v})}{q(\{\textbf p_n\}) q(\hat{\textbf v})}\right) \notag \\
     &= \sum_{n=1}^N - \mathbb E_{q(\hat{\textbf v})}\mathbb E_{ q(\textbf p_n)}\left[\ln p(\hat{\textbf x}_n | \textbf p_n, \hat{\textbf v}) \right] \notag \\
     &\qquad + \sum_{n=1}^N D_{\mathrm{KL}}\left[q(\textbf p_n)\|p(\textbf p_n)\right] 
    + D_{\mathrm{KL}}\left[q(\hat{\textbf v})\|p(\hat{\textbf v})\right].
   \label{eq:elbo_posevolume}
\end{align}}

The prior, approximate posterior and conditional likelihood all factorize across datapoints: $p(\{\textbf p_n\}) = \prod_{n=1}^N p(\textbf p_n)$, $q(\{\textbf p_n\}) = \prod_{n=1}^N q(\textbf p_n)$, and $p(\{\hat{\textbf x}_n\} | \{\textbf p_n\}, \hat{\textbf v}) = \prod_{n=1}^N p(\hat{\textbf x}_n | \textbf p_n, \hat{\textbf v})$. 
Like \eqref{eq:elbo_volume_minibatch}, for mini-batch optimization algorithms we make use of the objective
{\small
\begin{align}
    & \sum_{n\in D_i} -\mathbb E_{q(\hat{\textbf v})}\mathbb E_{ q(\textbf p_n)}\left[\ln p(\hat{\textbf x}_n | \textbf p_n, \hat{\textbf v}) \right]  \\
     &\quad + \sum_{n\in D_i} D_{\mathrm{KL}}\left[q(\textbf p_n)\|p(\textbf p_n)\right] 
    + \frac{|D_i|}{N} D_{\mathrm{KL}}\left[q(\hat{\textbf v})\|p(\hat{\textbf v})\right].\notag
\end{align}
}

In Section \ref{sec:VIstructure}, we learn the structure parameters shared across all data points. 
Here the pose parameters are unique per observation and therefore require separate inference procedures per data point. 
As an alternative, we can also learn  a function that approximates the inference procedure; 
This is called amortized inference \citep{kingma2013auto}. 
In practice, a complex parameterized function $f_{\phi}$ such as a neural network predicts the parameters of the variational posterior $\eta = f_{\phi}(\cdot)$.

\section{Experiments}

We empirically test the formulation above with simulated data from 
the well-known GroEL-GroES protein \citep{xu1997crystal}. 
To this end we generate three datasets, each with 40K projections
onto 128$\times$128 images at a resolution of 2.8\AA {} per pixel.
The three datasets have signal-to-noise ratios (SNR) of 1.0, 0.04 and 0.01, 
referred to below as the {\em noise-free}, {\em medium-noise} and {\em high-noise} cases. 
Figure \ref{fig:samples} shows one sample per noise level.
As previously stated in Section \ref{sec:image_formation}, we do not model the microscope's 
defocus or electron scattering effects, as captured by the CTF \citep{kohl2008transmission}.

Using synthetic data allows us to evaluate the algorithm with the  
ground-truth structure, e.g., in terms of mean-squared error (MSE). 
With real-data, where ground truth is unknown, the resolution of a fitted 3D structure 
is often quantified using {\em Fourier Shell Correlation} \citep{Rosenthal-Henderson-JMB2003}:
The $N$ observations are partitioned randomly into two sets, $A$ and $B$, 
each of which is then independently modeled with the same reconstruction algorithm. 
The normalized cross-correlation coefficient is then computed as a function of 
frequency  $f = 1/\lambda$ to assess the agreement between the two reconstructions.

Given two 3D structures,  $F_{A}$ and $F_{B}$,  in the Fourier domain,
FSC at frequency $f$ is given by
\begin{align}
 FSC(f|F_{A},F_{B})=
 \frac{ \sum\limits_{f_{i}\in S_f} F_{A}(f_{i})\cdot F_{B}(f_{i})^{\ast}}
 { \sqrt[{2}]{\sum\limits_{f_{i}\in S_f}{\left|F_{A}(f_{i})\right|^{2}}\cdot \sum\limits _{f_{i}\in S_f}{\left|F_{B}(f_{i})\right|^{2}}}  } ~.
\end{align}
where $S_f$ denotes the set of frequencies in a shell at distance $f$ from the 
origin of the Fourier domain (i.e. with wavelength $\lambda$).
This yields FSC curves like those in Fig. \ref{fig:FSC}. 
The quality (resolution) of the fit can be measured in terms of the frequency at 
which this curve crosses a threshold $\tau$. 
When one of $F_{A}$ or $F_{B}$ is ground truth, then $\tau=0.5$, and when both are
noisy reconstructions it is common to use $\tau=0.143$ \citep{Rosenthal-Henderson-JMB2003}).
The structure is then said to be resolved to wavelength $\lambda = 1/f$
for which $FSC(f) = \tau$. 

\subsection{Comparison to Baseline algorithms}
When poses are known, the current state-of-the-art (SOTA) is a conventional 
tomographic reconstruction (a.k.a.\ back-projection). 
When poses are unknown, there are several well-known SOTA cryo-EM algorithms \citep{de2013xmipp,grigorieff2007frealign,scheres2012relion,tang2007eman2}.
All provide point estimates of the 3D structure. 
In terms of graphical models, point estimates correspond to the case in which the posterior 
$q(\hat{\textbf{v}})$ is modeled as a delta function, $\delta(\hat{\textbf{v}} |\mu_{\mathbf{v}})$, 
the parameter of which is the 3D voxel array, $\mu_{\mathbf{v}}$.

We compare this baseline to a model in which the posterior  $q(\hat{\mathbf{v}})$ is a multivariate 
diagonal Gaussian, $\mathcal{N}(\hat{\textbf{v}}|\mu_{\textbf{v}},\mathbbm{1}\sigma_{\textbf{v}}^2)$. 
While the latent structure is modeled in Fourier domain, the spatial domain signal is real-valued.
We restrict the learnable parameters, $\mu_{\mathbf{v}}$ and $\sigma_{\mathbf{v}}$, accordingly
\footnote{That is $\forall \phi \in \{\mu_{\mathbf{v}},\sigma_{\mathbf{v}}\}$:
$\Re(\phi)[\zeta]=\Re(\phi)[-\zeta]$ and $\Im(\phi)[\zeta]=-\Im(\phi)[-\zeta]$  
with $\zeta=(\zeta_1,\zeta_2,\zeta_3)$.}. We use the reparameterization trick thus correlate the samples accordingly.
Finally, the prior in \eqref{eq:elbo_volume} in a multivariate standard Gaussian, 
$p( \mathbf{v}) = \mathcal{N}(\mathbf{v}|\mathbf{0}, \mathbbm{1})$. 

Table \ref{tab:exp1} shows results with these two models, with known poses
(the tomographic setting), and with {\em noise-free} observations.
Given the sensor resolution of $r=2.8\AA$, the highest possible resolution would be
the Nyquist wavelength of $5.6\AA$.
Our results show that both models approach this resolution, and in reasonable time.
\begin{table}[t]
\begin{center}
\begin{tabular}{c|rr}\toprule
 \multicolumn{1}{c}{\emph{Posterior}}  & \multicolumn{1}{c}{$\delta(\textbf{v}|\mu_{\mathbf{v}})$} & \multicolumn{1}{c}{$\mathcal{N}(\textbf{v}|\mu_{\mathbf{v}},\mathbbm{1}\sigma_{\mathbf{v}}^2)$} \\ \midrule
\emph{Time until}                & \multirow{2}{*}{$\sim$ 390}     & \multirow{2}{*}{$\sim$ 480}           \\
\emph{converged} ${[}s{]}$           &                              &                                    \\ \midrule
\emph{MSE}                         & \multirow{2}{*}{2.29}        & \multirow{2}{*}{2.62}              \\
{[}$10^{-3}$/voxel{]}       &                              &                                    \\ \midrule
\emph{Resolution} {[}\AA{]}                          & 5.82                         & 5.82    \\\bottomrule                                             
\end{tabular}
\end{center}
\caption{Results for modelling  protein structure as latent variable. Fitting a Gaussian or Dirac posterior distribution with VI leads to similar model fits, as measured by MSE and FSC between fit and ground truth with $\tau = 0.5$. }
\label{tab:exp1}
\end{table}
%

\subsection{Uncertainty estimation leads to data efficiency}
\label{sec:exp_data_eff}

\begin{figure}[thb]
  \begin{center}
    \includegraphics[width=0.48\textwidth]{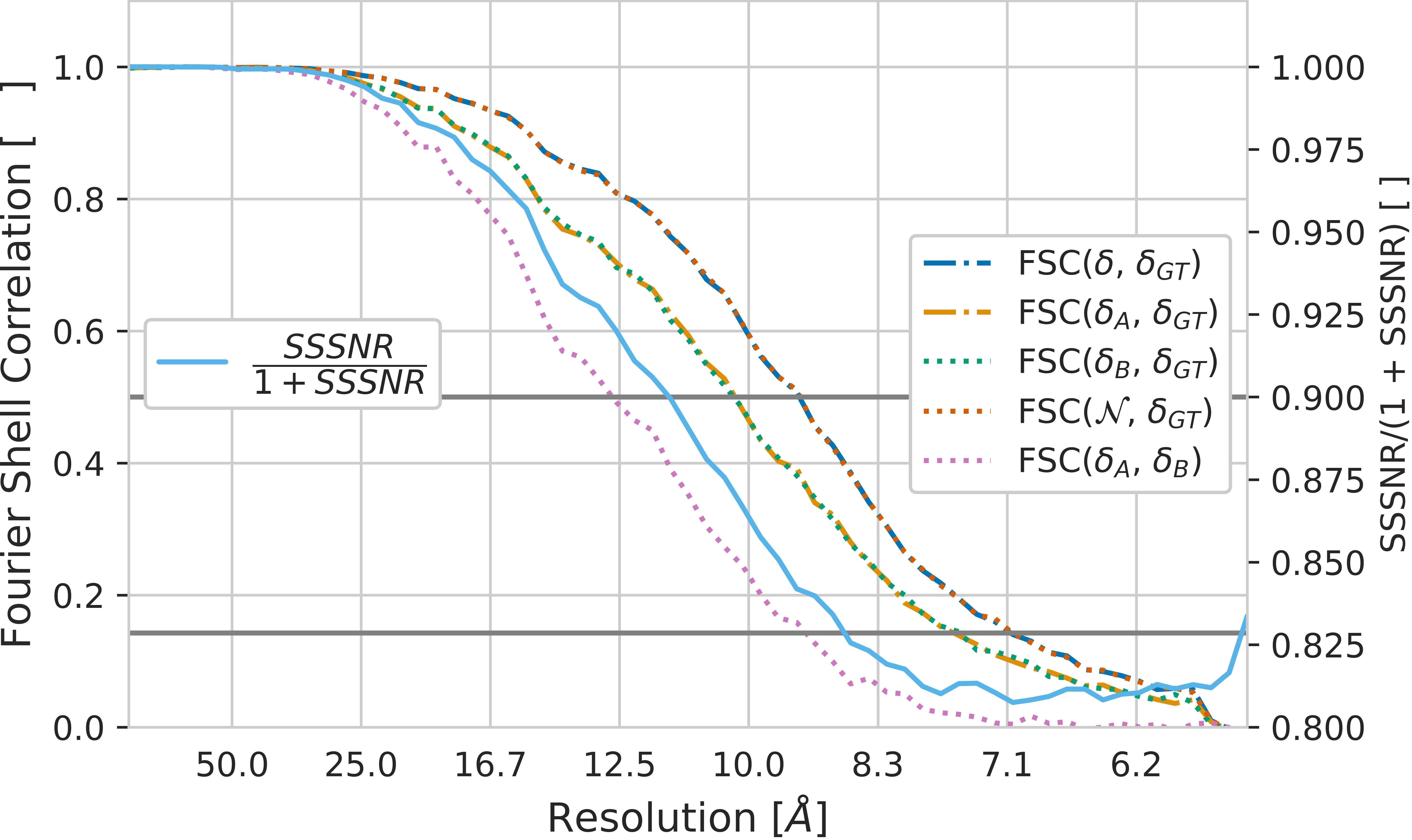} \\
    
    \vspace*{0.3cm}
    \includegraphics[width=0.48\textwidth]{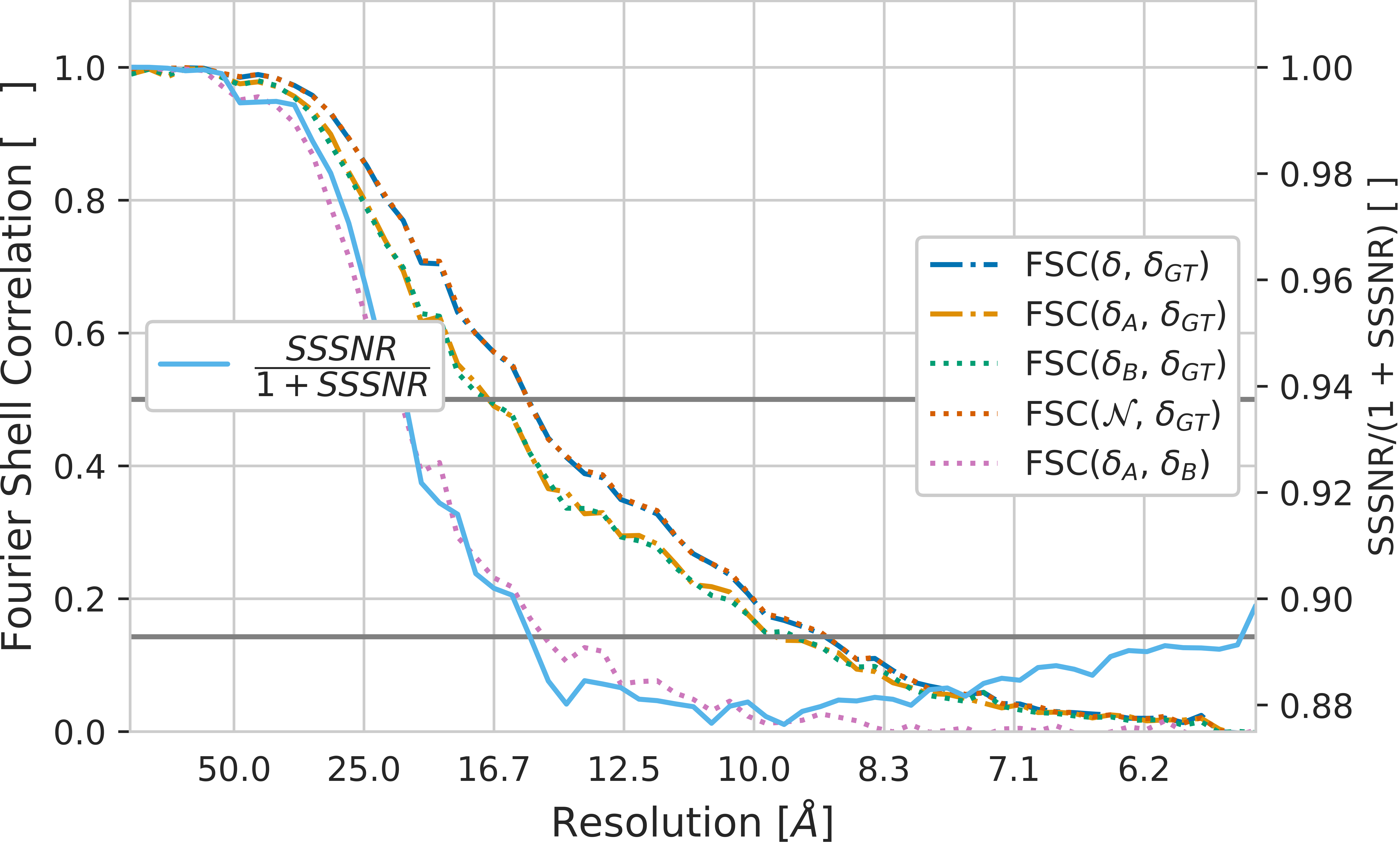}
  \end{center}
  \caption{The FSC curves for various model fits. 
  The grey lines indicate resolution thresholds. 
  The higher the resolution of a structure the better the model fit. 
  We contrast the FSC curves with the proposal we make to evaluate model fit.}
\label{fig:FSC}
\end{figure}

In this section, we explore how modelling the latent structure with uncertainty can improve data efficiency.
For this, recall that FSC is computed by comparing reconstructions based on dataset splits, $A$ and $B$.
As an alternative, we propose to utilize the learned model uncertainties $\sigma_{\mathbf{v}}$
to achieve a similar result. We thus only need one model fit that includes both dataset splits.
Specifically, we propose to extend a measure first presented in \citet{unser1987new}:
the spectral SNR to the 3D case, and hence  refer to it as spectral shell SNR (SS-SNR).
When modelling the latent structure as diagonal Gaussian $\mathcal{N}(\textbf{v}|\mu_{\mathbf{v}},\mathbbm{1}\sigma_{\mathbf{v}}^2)$, the SS-SNR can be computed to be
\begin{align}
    \alpha(f) =
    \frac{ \sum\limits_{f_{i}\in S_f} |\mu_{\mathbf{v}}(f_i)|^2}{\sum\limits_{f_{i}\in S_f}\sigma^2_{\mathbf{v}}(f_i)}.
\end{align}
Following the formulation by \cite{unser1987new}, we can then express the FSC
in terms of the SS-SNR, i.e., $FSC \approx \alpha / (1 + \alpha)$. 
%

Figure \ref{fig:FSC} shows  FSC curves based on reconstructions 
from the {\em medium-noise} (top) and {\em high-noise} (bottom) datasets.
First, we aim to demonstrate that the FSC curves between the Gaussian fit (with all data) 
vs ground truth, and the MAP estimate model with all data vs ground truth, i.e.,
FSC($f|\delta$, $\delta_{GT}$) and FSC($f|\mathcal{N} $, $\delta_{GT}$), yield the same fit quality. 
Note that we would not usually have access to the ground truth structure. 
Secondly, because in a realistic scenario we would not have access to the ground truth we would need to split the dataset in two. 
For this we evaluate the FSC between ground truth and two disjoint splits of the dataset  FSC($f|\delta_A$, $\delta_{GT}$) and FSC($f|\delta_B$, $\delta_{GT}$). 
This curve not surprisingly lies under the previous curves. Also note, that the actual measure we would consider FSC($f|\delta_A$, $\delta_B$) is more conservative. 
Finally, we show that $\alpha / (1 + \alpha)$  curve has the same inflection points as the FSC curve. 
As one would expect, it lies above the conservative FSC($f|\delta_A$, $\delta_B$). 

Using  $\alpha$ one can quantify the model fit with learned uncertainty rather than FSC curve.
As a consequence there is no need to partition the data and perform two 
separate reconstructions, each with only half the data.

\begin{figure}[thb]
  \begin{center}
    \includegraphics[width=0.45\textwidth]{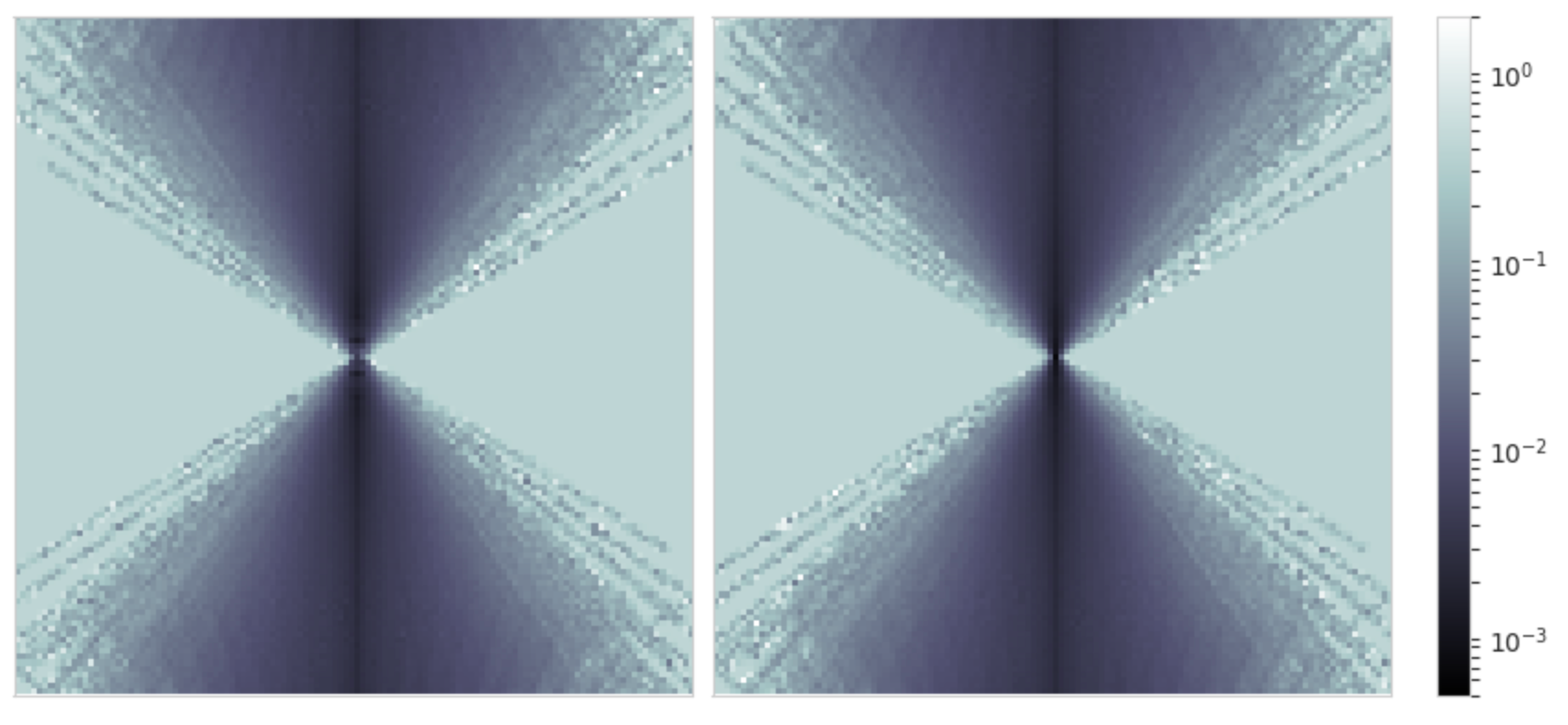}
  \end{center}
  \caption{Center slice through the learned Fourier volume uncertainties $\sigma_\mathbf{v}$. \textit{Left:} real part, \textit{Right:} imaginary part. 
  We learn the model fit with observations coming only from a $30^\circ$ cone, a scenario similar to breast cancer scans where observations are available only from some viewing directions. Uncertainty close to 1 means that the model has no information in these areas, close to zero represents areas of high sampling density.
  In contrast to other models, our model can identify precisely where information is missing (high variance).}
\label{fig:coneuncertainties}
\end{figure}

\subsection{Uncertainty identifies missing information}\label{sec:exp_cone}

Above we discussed how global uncertainty estimation can help estimate the quality of fit. 
Here we demonstrate how local uncertainty can help evaluate the local quality of 
fit\footnote{Other studies recognize the importance  of local uncertainty estimation, measuring
FSC locally in a wavelet basis \citep{cardone2013one,kucukelbir2014quantifying,vilas2018monores}.}. 
In many medical settings, such as breast cancer scans, limited viewing directions are available.
This is an issue in tomography, and also occurs in single particle cryo-EM 
when the distribution of particle orientations around the viewing sphere are highly anisotropic.
To recreate the same effect we construct a dataset of 40K observations, as before with no noise 
to separate the sources of information corruption\footnote{For completeness we present the same 
experiment with noise in the appendix.}.  
We restrict viewing directions to Euler angles $(\alpha,\beta,\gamma)$ with $\beta \in (-15°, +15°)$; 
i.e., no observations outside a $30^{\circ}$ cone 
As above, we assume a Gaussian posterior over the latent protein structure. 
\\
Figure \ref{fig:coneuncertainties} shows the result.
We can show that the uncertainty the model has learned correlates with regions in which data has been observed (uncertainty close to 0) and has not been observed (uncertainty close to 1). Due to the pressure from the KL-divergence (see \eqref{eq:elbo_volume}) the latter areas of the model default to the prior $\mathcal{N}(\mathbf{v}|\mathbf{0}, \mathbbm{1})$. 
This approach can be a helpful method to identify areas of low local quality of fit.


\subsection{Limits: Treating poses as random variables}

Extending our method from treating structures as latent variables to treating poses as latent variables is difficult. 
In the following we analyze why this is the case.
Note though that,  our method of estimating latent structure can be combined with common methods of pose estimation such as branch and bound \citep{punjani2017cryosparc} without losing any of the benefits we offer. 
However, ideally it would be interesting to also learn latent pose posteriors. 
This would be useful to for example detect outliers in the dataset which is common in real world scenarios. 

\begin{figure}[bt]
  \begin{center}
    \includegraphics[width=0.4\textwidth]{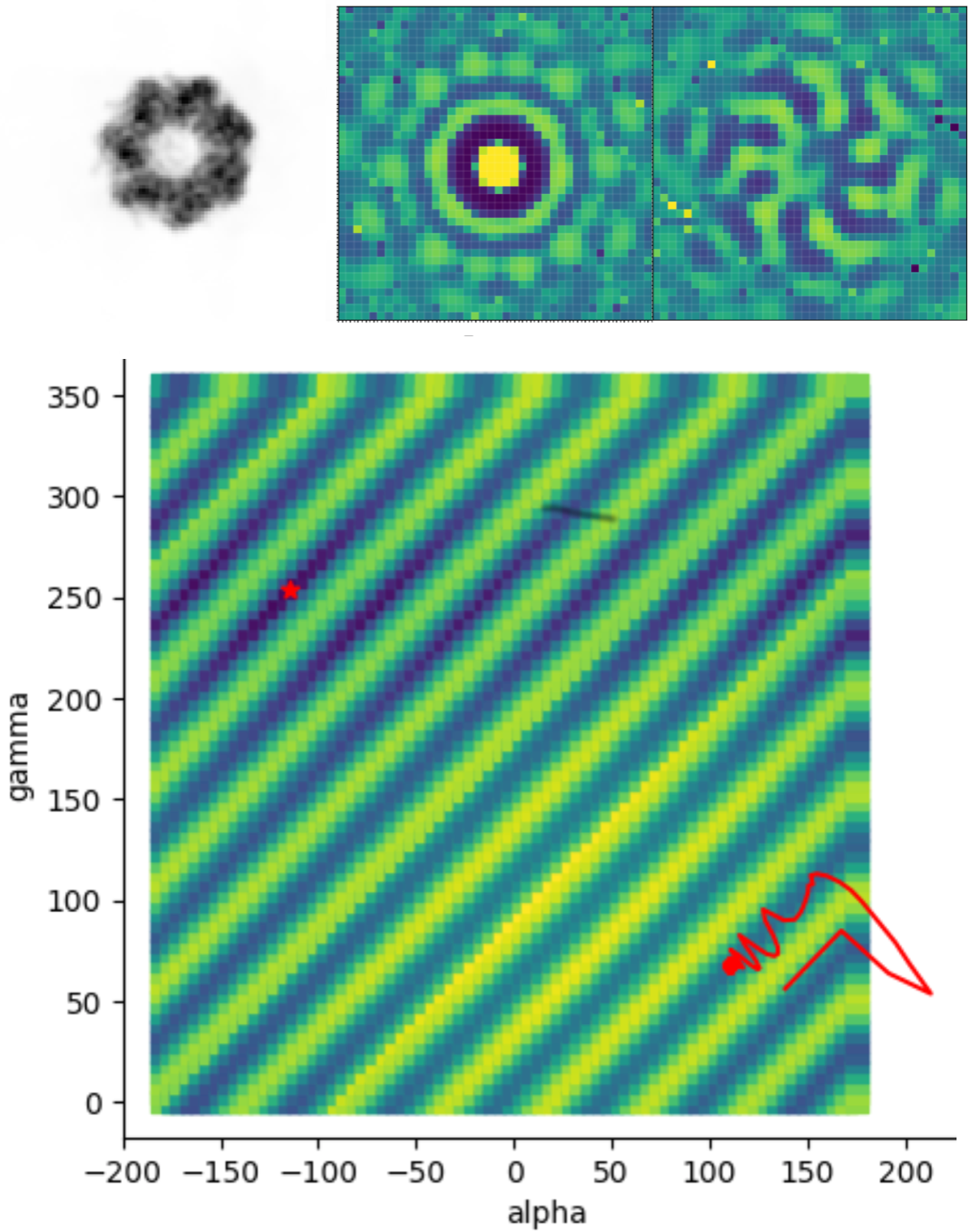}
  \end{center}
  \caption{Example of gradient descent failing to estimate the latent pose of a protein. \textit{Small images from left to right:} observation, real and imaginary part of the Fourier transpose. \textit{Large figure: } Corresponding error surface of the poses for 2 of 3 Euler angles. The red curve shows the progression of the pose estimate over the course of optimization. It is clear that the optimization fails to recover the true global optimum 
  (red star).}
\label{fig:fails}
\end{figure}

In an initial experiment, we fix the volume to the ground truth. We subsequently only estimate the poses with a simple Dirac model for each data point $\textbf{p}_n \sim \delta(\textbf{p}_n|\mu_{\mathbf{p}_n})$. 
In figure \ref{fig:fails}, we demonstrate the problem of SGD-based learning for this. For one example (samples on top), we show its true error surface, the global optimum (red star) and the  changes over the course of optimization (red line). We observe that due to the high symmetries in the structure, the true pose posterior error surface has many symmetries as well. An estimate depending on its starting position, seems to converge to the closest local optimum only rather than the global one. We would hope to be able to fix this problem in the future by applying more advanced density estimation approaches.

\section{Model Criticism and Future Work}

This paper introduces practical probabilistic models into the scientific imaging pipeline, where practical refers to scalability through the use of the reparameterization trick.
We show how to turn the operators in the pipeline into differentiable maps, as this is required to apply the trick.
The main focus of the experiments is to show why this novelty is important, addressing issues such as data efficiency, local uncertainty, and cross validation. 
Specifically, we found that a parameterized distribution, i.e. the Gaussian, achieves the same quality of fit as a point estimate, i.e. the dirac, while relying on less data. 
We conclude that our latent variable model is a suitable building block. It can be plugged into many SOTA approaches seamlessly, such as \citep{de2013xmipp,grigorieff2007frealign,scheres2012relion,tang2007eman2}.
We also established that the learned uncertainty is predictive of locations with too few samples.
Finally, we demonstrated the limits of our current methods in treating poses as latent variables.
This problem, however, does not limit the applicability of our method to latent structures. 
We thus propose to combine common pose estimation with our latent variable structure estimation. 
This method benefit from the uncertainty measure but also find globally optimal poses. 

In future work we hope to find a way to efficiently learn pose posterior distributions as well.
We hope that a reasonable approach would be to use multi-modal distributions and thus more advanced density estimation techniques. We will also try to incorporate amortized inference, mentioned in Section \ref{sec:VI}. 
Amortization would give the additional advantage of being able to transfer knowledge from protein to protein. 
Transfer could then lead to more advanced noise and CTF models.
Bias in transfer will be a key focus of this effort; i.e., we only want to transfer features of the noise and not the latent structure.
Another problem we see with the field of reconstruction algorithms is that the model evaluation can only help to detect variance but not bias in a model class. This is a problem with FSC comparison, but also with our proposal. We believe that an estimate of the data-log-likelihood of a hold out test dataset is generally much better suited. In a probabilistic view, this can be achieved by importance weighting the ELBO \citep{pmlr-v32-rezende14}. 

\newpage
\section*{Acknowledgements}
We thank the reviewers for their valuable feedback, in particular \textit{AnonReviewer1}. 
This research was funded in part by Google and by the Canadian Institute for Advanced Research.

\bibliography{references}

\begin{thebibliography}{}

\bibitem[Alemi et~al., 2017]{alemi2017fixing}
Alemi, A.~A., Poole, B., Fischer, I., Dillon, J.~V., Saurous, R.~A., and
  Murphy, K. (2017).
\newblock Fixing a broken elbo.
\newblock {\em arXiv preprint arXiv:1711.00464}.

\bibitem[Baxter et~al., 2009]{baxter2009determination}
Baxter, W.~T., Grassucci, R.~A., Gao, H., and Frank, J. (2009).
\newblock Determination of signal-to-noise ratios and spectral snrs in cryo-em
  low-dose imaging of molecules.
\newblock {\em Journal of structural biology}, 166(2):126--132.

\bibitem[Brubaker et~al., 2015]{brubaker2015building}
Brubaker, M.~A., Punjani, A., and Fleet, D.~J. (2015).
\newblock Building proteins in a day: Efficient 3d molecular reconstruction.
\newblock In {\em Proceedings of the IEEE Conference on Computer Vision and
  Pattern Recognition}, pages 3099--3108.

\bibitem[Cardone et~al., 2013]{cardone2013one}
Cardone, G., Heymann, J.~B., and Steven, A.~C. (2013).
\newblock One number does not fit all: Mapping local variations in resolution
  in cryo-em reconstructions.
\newblock {\em Journal of structural biology}, 184(2):226--236.

\bibitem[Chirikjian and Kyatkin, 2000]{chirikjian2000engineering}
Chirikjian, G. and Kyatkin, A. (2000).
\newblock {\em Engineering Applications of Noncommutative Harmonic Analysis:
  With Emphasis on Rotation and Motion Groups}.
\newblock CRC Press.

\bibitem[De~la Rosa-Trev{\'\i}n et~al., 2013]{de2013xmipp}
De~la Rosa-Trev{\'\i}n, J., Ot{\'o}n, J., Marabini, R., Zaldivar, A., Vargas,
  J., Carazo, J., and Sorzano, C. (2013).
\newblock Xmipp 3.0: an improved software suite for image processing in
  electron microscopy.
\newblock {\em Journal of structural biology}, 184(2):321--328.

\bibitem[Egelman, 2016]{egelman2016current}
Egelman, E.~H. (2016).
\newblock The current revolution in cryo-em.
\newblock {\em Biophysical journal}, 110(5):1008--1012.

\bibitem[Faruqi et~al., 2003]{faruqi2003evaluation}
Faruqi, A., Cattermole, D., Henderson, R., Mikulec, B., and Raeburn, C. (2003).
\newblock Evaluation of a hybrid pixel detector for electron microscopy.
\newblock {\em Ultramicroscopy}, 94(3-4):263--276.

\bibitem[Grigorieff, 1998]{grigorieff1998three}
Grigorieff, N. (1998).
\newblock Three-dimensional structure of bovine nadh: ubiquinone oxidoreductase
  (complex i) at 22 {\aa} in ice.
\newblock {\em Journal of molecular biology}, 277(5):1033--1046.

\bibitem[Grigorieff, 2007]{grigorieff2007frealign}
Grigorieff, N. (2007).
\newblock Frealign: high-resolution refinement of single particle structures.
\newblock {\em Journal of structural biology}, 157(1):117--125.

\bibitem[Havin and Jöricke, 1994]{uncertainty_principle}
Havin, V. and Jöricke, B. (1994).
\newblock {\em The Uncertainty Principle in Harmonic Analysis}.
\newblock Springer-Verlag.

\bibitem[Henderson et~al., 2012]{henderson2012outcome}
Henderson, R., Sali, A., Baker, M.~L., Carragher, B., Devkota, B., Downing,
  K.~H., Egelman, E.~H., Feng, Z., Frank, J., Grigorieff, N., et~al. (2012).
\newblock Outcome of the first electron microscopy validation task force
  meeting.
\newblock {\em Structure}, 20(2):205--214.

\bibitem[Jaderberg et~al., 2015]{jaderberg2015spatial}
Jaderberg, M., Simonyan, K., Zisserman, A., et~al. (2015).
\newblock Spatial transformer networks.
\newblock In {\em Advances in Neural Information Processing Systems}, pages
  2017--2025.

\bibitem[Jaitly et~al., 2010]{jaitly2010bayesian}
Jaitly, N., Brubaker, M.~A., Rubinstein, J.~L., and Lilien, R.~H. (2010).
\newblock A bayesian method for 3d macromolecular structure inference using
  class average images from single particle electron microscopy.
\newblock {\em Bioinformatics}, 26(19):2406--2415.

\bibitem[Jordan et~al., 1998]{jordan1998introduction}
Jordan, M.~I., Ghahramani, Z., Jaakkola, T.~S., and Saul, L.~K. (1998).
\newblock An introduction to variational methods for graphical models.
\newblock In {\em Learning in graphical models}, pages 105--161. Springer.

\bibitem[Kingma and Welling, 2013]{kingma2013auto}
Kingma, D.~P. and Welling, M. (2013).
\newblock Auto-encoding variational bayes.
\newblock {\em arXiv preprint arXiv:1312.6114}.

\bibitem[Kohl and Reimer, 2008]{kohl2008transmission}
Kohl, H. and Reimer, L. (2008).
\newblock {\em Transmission electron microscopy: physics of image formation}.
\newblock Springer.

\bibitem[Kucukelbir et~al., 2014]{kucukelbir2014quantifying}
Kucukelbir, A., Sigworth, F.~J., and Tagare, H.~D. (2014).
\newblock Quantifying the local resolution of cryo-em density maps.
\newblock {\em Nature methods}, 11(1):63.

\bibitem[Langlois et~al., 2014]{langlois2014automated}
Langlois, R., Pallesen, J., Ash, J.~T., Ho, D.~N., Rubinstein, J.~L., and
  Frank, J. (2014).
\newblock Automated particle picking for low-contrast macromolecules in
  cryo-electron microscopy.
\newblock {\em Journal of structural biology}, 186(1):1--7.

\bibitem[Marino et~al., 2018]{marino2018iterative}
Marino, J., Yue, Y., and Mandt, S. (2018).
\newblock Iterative amortized inference.
\newblock {\em arXiv preprint arXiv:1807.09356}.

\bibitem[Pettersen et~al., 2004]{pettersen2004ucsf}
Pettersen, E.~F., Goddard, T.~D., Huang, C.~C., Couch, G.~S., Greenblatt,
  D.~M., Meng, E.~C., and Ferrin, T.~E. (2004).
\newblock Ucsf chimera—a visualization system for exploratory research and
  analysis.
\newblock {\em Journal of computational chemistry}, 25(13):1605--1612.

\bibitem[Pintilie, 2010]{Pintilie}
Pintilie, G. (2010).
\newblock Greg pintilie's homepage:
  \texttt{people.csail.mit.edu/gdp/cryoem.html}.

\bibitem[Punjani et~al., 2017]{punjani2017cryosparc}
Punjani, A., Rubinstein, J.~L., Fleet, D.~J., and Brubaker, M.~A. (2017).
\newblock cryosparc: algorithms for rapid unsupervised cryo-em structure
  determination.
\newblock {\em Nature Methods}, 14(3):290--296.

\bibitem[Rezende et~al., 2016]{rezende2016unsupervised}
Rezende, D.~J., Eslami, S.~A., Mohamed, S., Battaglia, P., Jaderberg, M., and
  Heess, N. (2016).
\newblock Unsupervised learning of 3d structure from images.
\newblock In {\em Advances in Neural Information Processing Systems}, pages
  4996--5004.

\bibitem[Rezende et~al., 2014a]{rezende2014stochastic}
Rezende, D.~J., Mohamed, S., and Wierstra, D. (2014a).
\newblock Stochastic backpropagation and approximate inference in deep
  generative models.
\newblock {\em arXiv preprint arXiv:1401.4082}.

\bibitem[Rezende et~al., 2014b]{pmlr-v32-rezende14}
Rezende, D.~J., Mohamed, S., and Wierstra, D. (2014b).
\newblock Stochastic backpropagation and approximate inference in deep
  generative models.
\newblock In {\em Proceedings of the 31st International Conference on Machine
  Learning}, Proceedings of Machine Learning Research, pages 1278--1286.

\bibitem[Rosenthal and Henderson, 2003]{Rosenthal-Henderson-JMB2003}
Rosenthal, P.~B. and Henderson, R. (2003).
\newblock Optimal determination of particle orientation, absolute hand, and
  contrast loss in single-particle electron cryomicroscopy.
\newblock {\em Journal of Molecular Biology}, 333(4):721--745.

\bibitem[Rupp, 2009]{rupp2009biomolecular}
Rupp, B. (2009).
\newblock {\em Biomolecular crystallography: principles, practice, and
  application to structural biology}.
\newblock Garland Science.

\bibitem[Scheres, 2012]{scheres2012relion}
Scheres, S.~H. (2012).
\newblock Relion: implementation of a bayesian approach to cryo-em structure
  determination.
\newblock {\em Journal of structural biology}, 180(3):519--530.

\bibitem[Scheres et~al., 2007a]{scheres2007disentangling}
Scheres, S.~H., Gao, H., Valle, M., Herman, G.~T., Eggermont, P.~P., Frank, J.,
  and Carazo, J.-M. (2007a).
\newblock Disentangling conformational states of macromolecules in 3d-em
  through likelihood optimization.
\newblock {\em Nature methods}, 4(1):27.

\bibitem[Scheres et~al., 2007b]{scheres2007modeling}
Scheres, S.~H., N{\'u}{\~n}ez-Ram{\'\i}rez, R., G{\'o}mez-Llorente, Y.,
  San~Mart{\'\i}n, C., Eggermont, P.~P., and Carazo, J.~M. (2007b).
\newblock Modeling experimental image formation for likelihood-based
  classification of electron microscopy data.
\newblock {\em Structure}, 15(10):1167--1177.

\bibitem[Sigworth, 1998]{sigworth1998maximum}
Sigworth, F. (1998).
\newblock A maximum-likelihood approach to single-particle image refinement.
\newblock {\em Journal of structural biology}, 122(3):328--339.

\bibitem[Sigworth, 2016]{sigworth2016principles}
Sigworth, F.~J. (2016).
\newblock Principles of cryo-em single-particle image processing.
\newblock {\em Microscopy}, 65(1):57--67.

\bibitem[Tang et~al., 2007]{tang2007eman2}
Tang, G., Peng, L., Baldwin, P.~R., Mann, D.~S., Jiang, W., Rees, I., and
  Ludtke, S.~J. (2007).
\newblock Eman2: an extensible image processing suite for electron microscopy.
\newblock {\em Journal of structural biology}, 157(1):38--46.

\bibitem[Unser et~al., 1987]{unser1987new}
Unser, M., Trus, B.~L., and Steven, A.~C. (1987).
\newblock A new resolution criterion based on spectral signal-to-noise ratios.
\newblock {\em Ultramicroscopy}, 23(1):39--51.

\bibitem[Vilas et~al., 2018]{vilas2018monores}
Vilas, J.~L., G{\'o}mez-Blanco, J., Conesa, P., Melero, R., de~la
  Rosa-Trev{\'\i}n, J.~M., Ot{\'o}n, J., Cuenca, J., Marabini, R., Carazo,
  J.~M., Vargas, J., et~al. (2018).
\newblock Monores: automatic and accurate estimation of local resolution for
  electron microscopy maps.
\newblock {\em Structure}, 26(2):337--344.

\bibitem[Vulovi{\'c} et~al., 2013]{vulovic2013image}
Vulovi{\'c}, M., Ravelli, R.~B., van Vliet, L.~J., Koster, A.~J., Lazi{\'c},
  I., L{\"u}cken, U., Rullg{\aa}rd, H., {\"O}ktem, O., and Rieger, B. (2013).
\newblock Image formation modeling in cryo-electron microscopy.
\newblock {\em Journal of structural biology}, 183(1):19--32.

\bibitem[Xu et~al., 1997]{xu1997crystal}
Xu, Z., Horwich, A.~L., and Sigler, P.~B. (1997).
\newblock The crystal structure of the asymmetric groel--groes--(adp) 7
  chaperonin complex.
\newblock {\em Nature}, 388(6644):741.

\bibitem[Zhao et~al., 2013]{zhao2013tmacs}
Zhao, J., Brubaker, M.~A., and Rubinstein, J.~L. (2013).
\newblock Tmacs: A hybrid template matching and classification system for
  partially-automated particle selection.
\newblock {\em Journal of structural biology}, 181(3):234--242.

\end{thebibliography}
\bibliographystyle{apalike}

\newpage
\clearpage
\begin{appendices}
\section{Visual impressions}
\begin{figure}[htb]
  \begin{center}
    \includegraphics[width=0.45\textwidth]{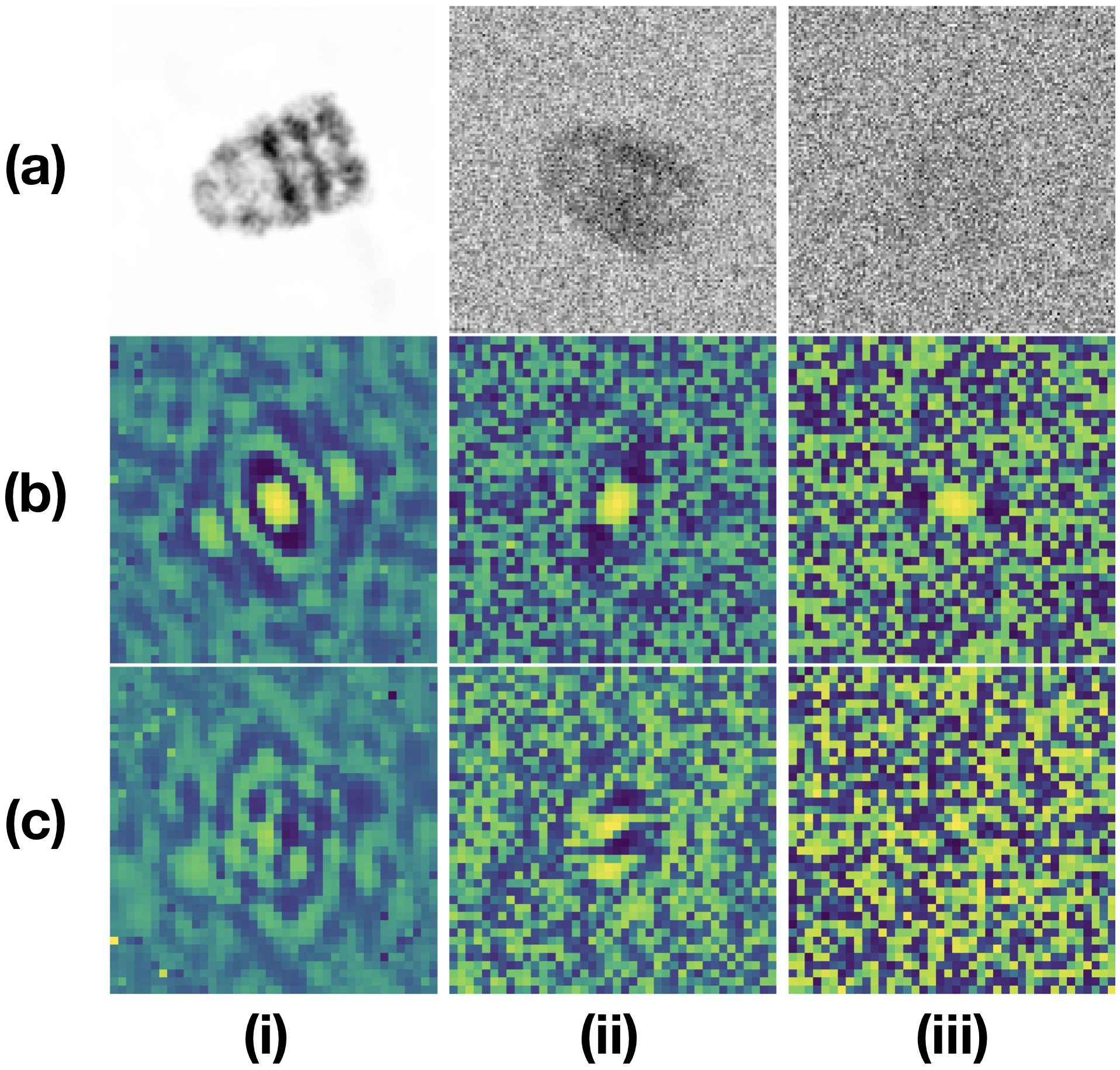}
  \end{center}
  \caption{\textit{Left to Right:} We show samples from the dataset we use: (i) no noise (such as in Experiment 6.1), (ii) moderate noise and (iii) high noise (such as in experiment 6.2). \textit{Top to Bottom:} Observation in (a) real space, first 20 Fourier shells (b) real part and (c) imaginary part (for better visibility log-scaled). The latter two are being used for the optimization due to the application of the Fourier slice theorem explained in Section \ref{sec:fourier_slice}.}
\label{fig:samples}
\end{figure}

\begin{figure}[htb]
  \begin{center}
    \includegraphics[width=0.45\textwidth]{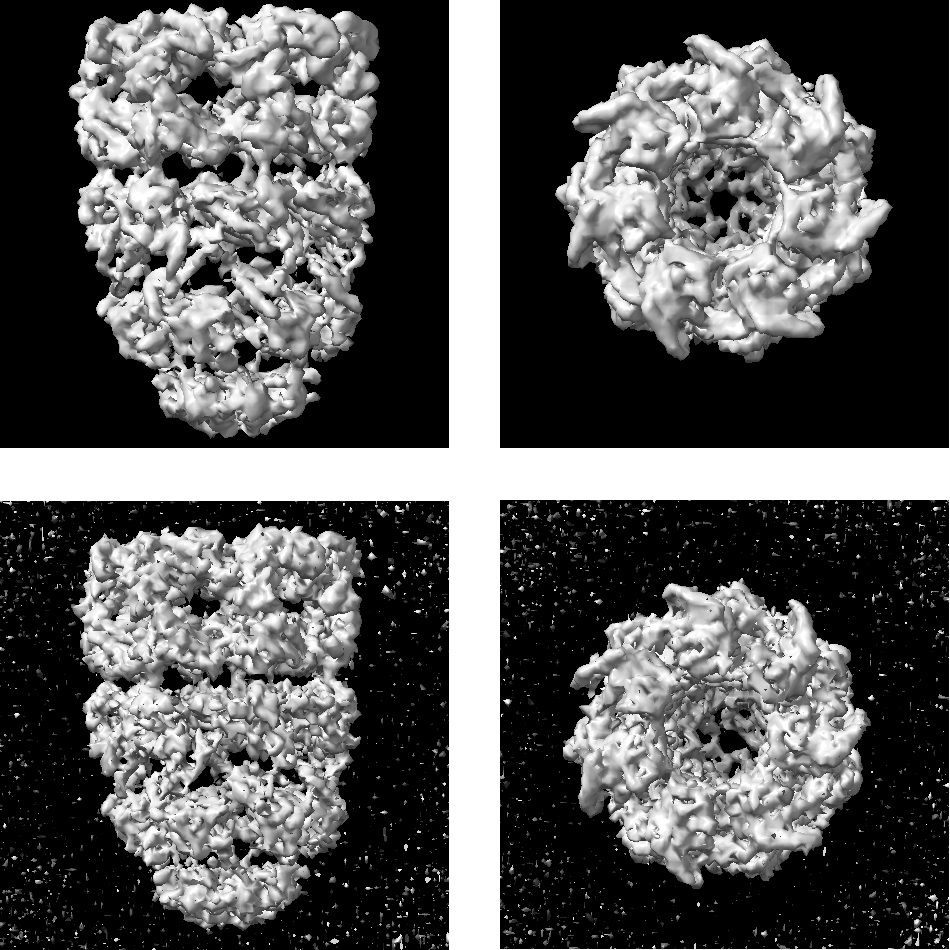}
  \end{center}
  \caption{\textit{Top:} Side and top view of the GroEL-GroES protein fit with moderate noise level data.
  \textit{Bottom:} Side and top view of the respective high noise level dataset.}
\label{fig:reconstructions}
\end{figure}

First, to give a visual impression of the observations, we learn from we present in Figure \ref{fig:samples} samples from the 3 data sets we use. All of the datasets are based on the same protein estimate of the GroEL-GroES protein\citep{xu1997crystal}. They differ in the signal-to-noise ratio (SNR). The left row represents noise free data, the middle a moderate common noise level, and the right an extreme level of noise.
For each observation example, we show also the corresponding Fourier transformations, their real part in the second column and their imaginary part in the third column.
Further we present the qualitative results of fitting the mid and high level datasets with our method. We visualize the respective protein fit from experiment 6.2 in Figure \ref{fig:reconstructions} with the Chimera X software package \citep{pettersen2004ucsf}. The two pictures on the top row represent the middle noise fit, respectively the bottom two the high noise fit.

\section{Extension to experiment \ref{sec:exp_cone}}

\begin{figure}[htb]
  \begin{center}
    \includegraphics[width=0.45\textwidth]{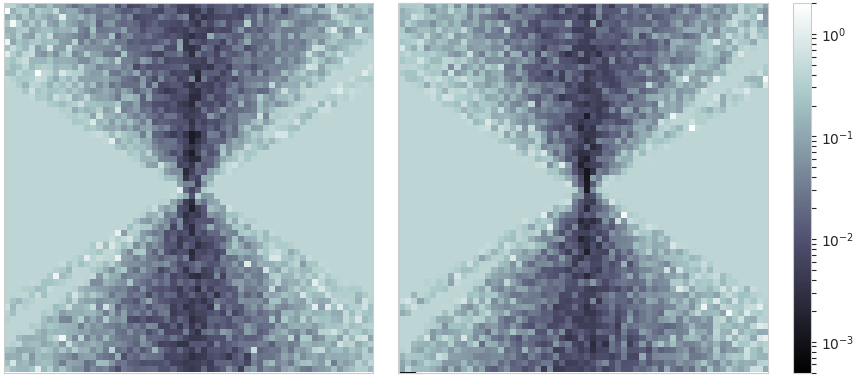}
  \end{center}
  \caption{Center slice through the learned Fourier volume uncertainties $\sigma_\mathbf{v}$. \textit{Left:} real part, \textit{Right:} imaginary part. 
  We learn the model fit with observations coming only from a $30^\circ$ cone, a scenario similar to breast cancer scans where observations are available only from some viewing directions. Uncertainty close to 1 means that the model has no information in these areas, close to zero represents areas of high sampling density.
  In contrast to other models, our model can identify precisely where information is missing (high variance).}
\label{fig:noisy_cone}
\end{figure}

\begin{figure}[htb]
  \begin{center}
    \includegraphics[width=0.45\textwidth]{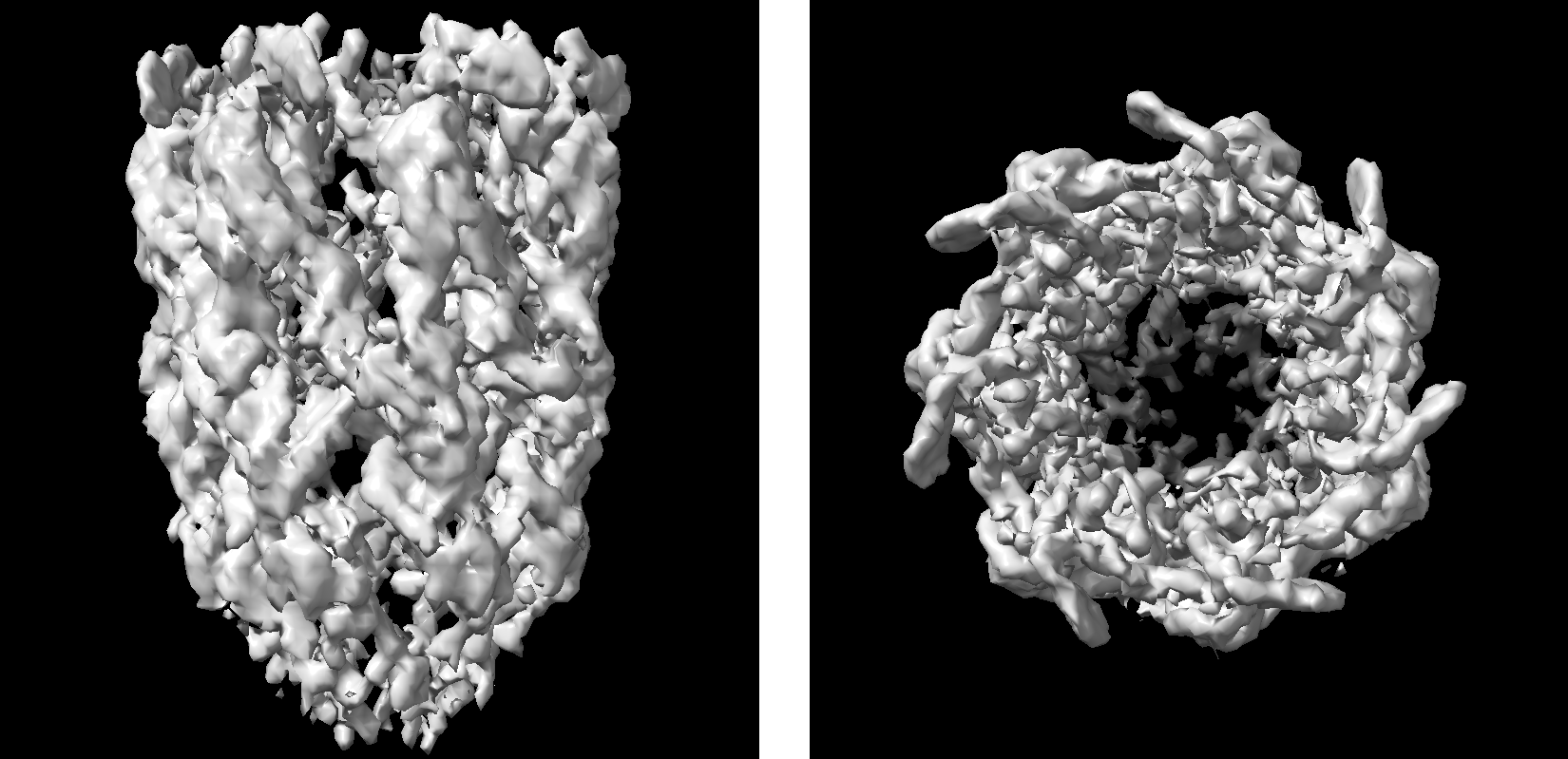}
  \end{center}
  \caption{Side and top view of the GroEL-GroES protein fit with moderate noise level data and all observations stemming from a limited pose space.}
\label{fig:cone_reconstructions}
\end{figure}

We shall execute the same experiment as in section \ref{sec:exp_cone} given the dataset with intermediate noise.
We display the experiments of this result in Figure \ref{fig:noisy_cone}. It is clear that while, missing information leads to large deviation in variance we also find that the noise leads to some variance in the observed area.
Again we visualize the result of the fit in Figure \ref{fig:cone_reconstructions}.

\section{Amortized inference and variational EM for pose estimation}

We have not used amortized inference in our experiments. In experiment 6.4 we have modelled poses as local variables and trained them by variational expectation maximization. Other work shows that the amortization gab can be significant \citep{scheres2007modeling,alemi2017fixing,marino2018iterative}. Hence in order to exclude the gap as a reason for failure, we decided to model local variables. We did run experiments though with ResNets as encoders without success either. We believe the core problem in probabilistic pose estimation is the number of local optima. This makes simple SGD a somewhat poor choice, because we rely on finding the global minimum. 

\section{Remarks to the chosen observation model}
The Gaussian noise model is a common but surely oversimplified model \citep{sigworth1998maximum}. A better model would be the Poisson distribution. The Gaussian is a good approximation to it given a high rate parameter meaning if there is a reasonable high count of radiation hitting the sensors. This is a good assumption for most methods of the field, but can actually be a poor model in some cases of cryo electron microscopy. An example of an elaborate model is presented in \citet{vulovic2013image}. 

\end{appendices}
\end{document}